\documentclass[letterpaper, 10 pt, conference]{ieeeconf}  

\IEEEoverridecommandlockouts                              

\usepackage{graphics} 
\usepackage{epsfig} 
\usepackage{textcomp}
\usepackage{stfloats}
\usepackage{url}
\usepackage{verbatim}
\usepackage{graphicx}
\usepackage{cite}
\usepackage{tikz}
\usepackage{comment}
\usepackage{amsmath,amssymb} 
\usepackage{color}

\usepackage{wasysym}
\usepackage{graphics} 
\usepackage{epsfig} 
\usepackage{mathptmx} 
\usepackage{tikz}
\usepackage{comment}
\usepackage{color}

\usepackage{booktabs}
\usepackage{wasysym}
\usepackage{graphics} 
\usepackage{epsfig} 
\usepackage{mathptmx} 
\usepackage{color} 
\usepackage{tabularx}
\usepackage{makecell}
\usepackage{multicol}
\usepackage{mwe,lipsum}
\usepackage{array,multirow}
\usepackage{float}
\usepackage{eucal}
\usepackage{subcaption}
\usepackage{xcolor}
\usepackage{nicematrix}
\usepackage{color, colortbl}
\definecolor{LightCyan}{rgb}{0.88,1,1}

\usepackage{hyperref}

\title{\LARGE \bf
RGB-Event Fusion for Moving Object Detection in Autonomous Driving
}

\author{Zhuyun Zhou$^{1}$, Zongwei Wu$^{*1,2}$, Rémi Boutteau$^{3}$, Fan Yang$^{1}$, Cédric Demonceaux$^{1,4}$, Dominique Ginhac$^{1}$
\thanks{This work was supported by the French National Research Agency through ANR CERBERE (ANR-21-CE22-0006).}
\thanks{$^{1}$Zhuyun Zhou, Zongwei Wu, Fan Yang, Cédric Demonceaux, and Dominique Ginhac are with ImViA, University of Burgundy (Université de Bourgogne), Dijon, France. {\tt\small \{Zhuyun\_Zhou@etu., Zongwei\_Wu@etu., fanyang@, Cedric.Demonceaux@, dginhac@\} u-bourgogne.fr }}%
\thanks{$^{2}$Zongwei Wu is also with CVL, ETH Zurich.}
\thanks{$^{3}$Rémi Boutteau is with Univ Rouen Normandie, LITIS, UR 4108, F-76000 Rouen, France. {\tt\small remi.boutteau@univ-rouen.fr}}
\thanks{$^{4}$Cédric Demonceaux is also with Université de Lorraine, CNRS, Inria, LORIA, Nancy, France.}
\thanks{$^{*}$Corresponding author: {\tt\small zongwei.wu.97@gmail.com}}
}

\begin{document}

\maketitle
\thispagestyle{empty}
\pagestyle{empty}

\begin{abstract}

Moving Object Detection (MOD) is a critical vision task for successfully achieving safe autonomous driving. Despite plausible results of deep learning methods, most existing approaches are only frame-based and may fail to reach reasonable performance when dealing with dynamic traffic participants. Recent advances in sensor technologies, especially the Event camera, can naturally complement the conventional camera approach to better model moving objects. However, event-based works often adopt a pre-defined time window for event representation, and simply integrate it to estimate image intensities from events, neglecting much of the rich temporal information from the available asynchronous events. Therefore, from a new perspective, we propose RENet, a novel RGB-Event fusion Network, that jointly exploits the two complementary modalities to achieve more robust MOD under challenging scenarios for autonomous driving. Specifically, we first design a temporal multi-scale aggregation module to fully leverage event frames from both the RGB exposure time and larger intervals. Then we introduce a bi-directional fusion module to attentively calibrate and fuse multi-modal features. To evaluate the performance of our network, we carefully select and annotate a sub-MOD dataset from the commonly used DSEC dataset. Extensive experiments demonstrate that our proposed method performs significantly better than the state-of-the-art RGB-Event fusion alternatives. The source code and dataset are publicly available at: \href{https://github.com/ZZY-Zhou/RENet}{https://github.com/ZZY-Zhou/RENet}.

\end{abstract}

\section{INTRODUCTION}

With the development of deep learning methods, researchers have reported great success in modeling the motion behaviors of different traffic participants \cite{siam2018modnet, rashed2019fusemodnet, Baur_2021_ICCV}.
However, most existing methods are frame-based and their performances are highly dependent on video quality, i.e., the frame rate and the image quality \cite{mackin2018study, madhusudana2021st,sun2022coarse}.
The former (frame rate) is essential to model the temporal consistency and the motion behavior, while the latter (image quality) influences the accuracy of object detection within a single frame. Therefore, despite significant advances in deep networks, the inherent limits of frame-based cameras are one of the main performance bottlenecks for moving object detection.

Recent advances in bio-inspired event cameras have drawn great research attention for autonomous vehicles \cite{chen2020event, li2019event,maqueda2018event}.
The rich temporal information provided by events can help to better avoid collisions with moving traffic agents and thus yield safer autonomous driving. 
In fact, when dealing with complex lighting scenes, such as driving during night or passing through a tunnel, the performance of frame-based methods may degrade severely \cite{zhen2019estimating,wu2021modality,wu2022robust,lin2022r}. Meanwhile, event cameras are more robust in these adverse visual conditions since they can quickly adapt to light changes thanks to its low latency. Nevertheless, event data often provide wider object boundaries than RGB frames. This is mainly due to the synchronization between RGB and events since RGB frames are obtained during a short exposure time, while event cameras work in an asynchronous mode at a higher temporal resolution. Therefore, how to fully leverage the heterogeneous but mutually complementary modalities is yet an open research topic.


\begin{figure}[t]
\centering
\includegraphics[width=\linewidth]{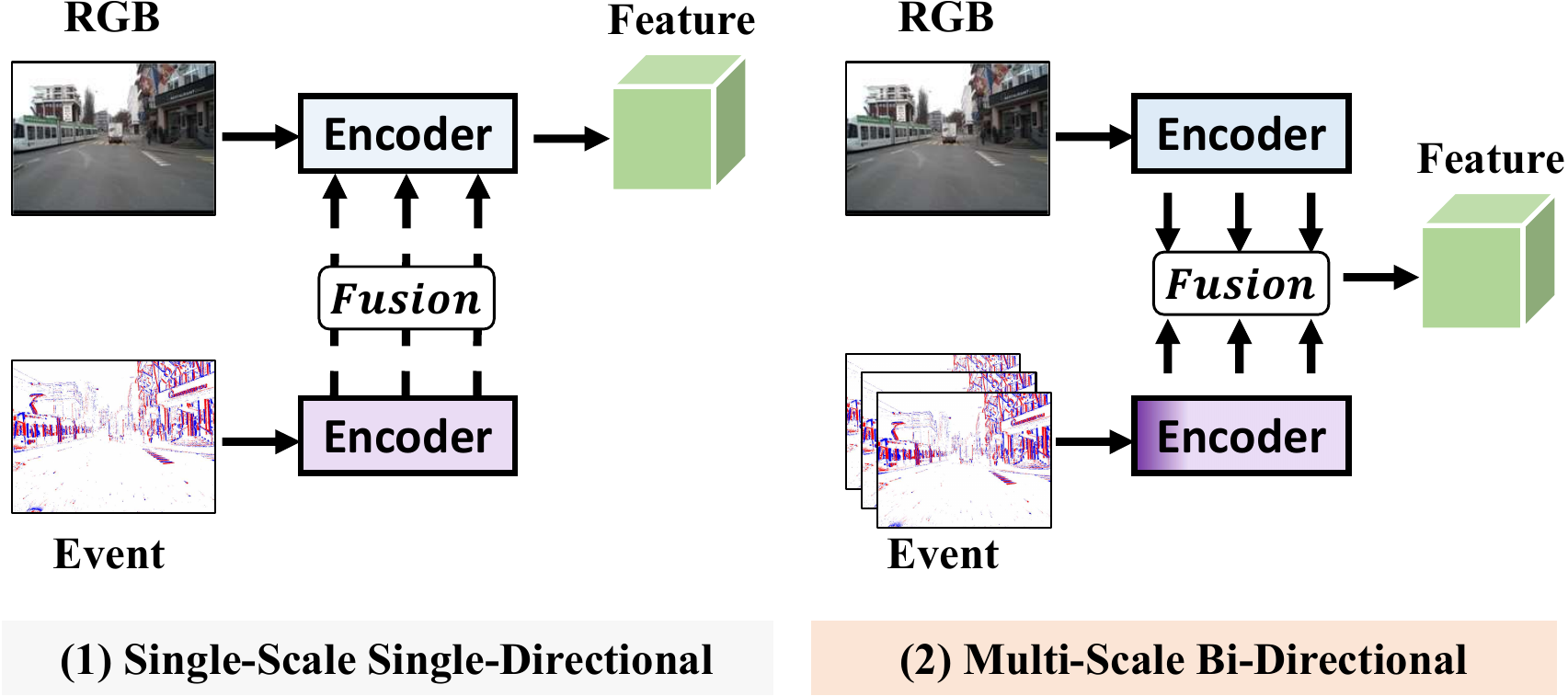}
\caption{\textbf{Comparison with RGB-Event Fusion Methods.} (1) Most existing works take a single-range event as input \cite{Messikommer2022bridging,tomy2022fusing,Tulyakov2022timlens++,gehrig2021raft} with a single-directional fusion design \cite{tomy2022fusing,sun2022event}, which neglects the rich temporal cues and cannot thoroughly explore the cross-modal relationship. (2) Our proposed network can benefit from multi-scale temporal events to better deal with object motion. We further adopt a bi-directional fusion design to better model multi-modal features and attentively form the shared representation.}
\label{Fig-intro}
\vspace{-3mm}
\end{figure}

In the literature, a common practice \cite{sun2022event,tulyakov2021time, tulyakov2022time, tomy2022fusing} is to gradually merge event features in the RGB stream to guide the feature modeling as shown in Fig. \ref{Fig-intro}.
Despite the achieved plausible results, these works are often uni-directional and cannot thoroughly leverage modality-specific cues during the encoding stages, i.e., RGB-only, event-only, and fused features.
Moreover, most existing works choose a pre-defined time interval as a hyper-parameter for event representation, which yields two major shortages: first, the fixed event representation cannot adapt to different scenarios without handcraft adjusting; secondly, a small range of event intervals, such as the exposure time of RGB camera, cannot fully leverage the rich temporal cues from the event data, while a long range of event intervals adds significant misaligned response.

To address the aforementioned issues, in this paper, we propose a novel RGB-Event network, named RENet, for moving object detection.
Different from previous works basing on single-range events, as shown in Fig. \ref{Fig-intro}, our event stream takes multi-range events as inputs to better benefit from the rich temporal cues.
Specifically, for each event representation, we first apply convolutions to project the input event maps to a latent space.
We then apply pooling with different kernel sizes to deal with events from different time ranges, i.e., larger pooling windows for longer-range events.
Our motivation comes from the observation that the event frames of long-time ranges introduce local motion blur around the object boundaries.
The motion blurs are essential for the detection of moving objects, while being not well-aligned with the image frame obtained during the exposure time.
Therefore, we apply different sizes of max-pooling to preserve the most informative motion features and improve the scale invariance within a local region, yielding a simple yet efficient way to aggregate rich temporal events in a coarse-to-fine manner.

Once the multi-range events are fused in the latent space, we feed them through the dual-residual networks along with RGB features. To attentively integrate multi-modal features, we introduce a bi-directional calibration module that first improves each input feature by attending to the channel and the spatial dimensions of the input features and then realizes a cross-modal calibration.
Finally, the cross-calibrated features are merged and further integrated with the hierarchical features, if any, to form the shared output. To the best of our knowledge, we are the first to apply attention modules for RGB-Event (moving) object detection tasks. Additionally, we provide DSEC-MOD, a new dataset dedicated to moving object detection from the widely used RGB-Event dataset DSEC \cite{gehrig2021dsec}. DSEC-MOD contains moving objects with automatically labeled and manually checked bounding boxes. Extensive comparisons performed on DSEC-MOD demonstrate the effectiveness of RENet against state-of-the-art RGB-Event fusion alternatives.

To summarize, our main contributions are three-fold:
\begin{itemize}

\item We design a multi-range aggregation module to fully leverage the rich temporal information from events, which are crucial for moving object detection.

\item We introduce a novel middle fusion design for RGB-Event fusion that models both modality-specific and shared representations.
\item We propose a novel DSEC-MOD dataset with automatically labeled and manually verified annotation to encourage the development of moving object detection with events. We benchmark various state-of-the-art (SOTA) fusion modules on our dataset and show that RENet significantly outperforms the fusion alternatives. 

\end{itemize}

\section{RELATED WORK}

\noindent\textbf{Event Datasets:} Unlike traditional datasets for global shutter RGB cameras, the number of available public event datasets is much fewer, and the types of tasks that may be practiced with these datasets are also limited. For example, several datasets \cite{hu2021v2e, gehrig2021dsec, hidalgo2020learning, zhu2018multivehicle}  only provide depth information as ground truth, including the two most widely used DSEC \cite{gehrig2021dsec} and MVSEC \cite{zhu2018multivehicle}. To encourage the development of segmentation and detection tasks for autonomous driving, recent researches propose to induce from these two datasets and create sub-datasets for specific tasks. For example, MVSEC-NIGHT \cite{hu2021v2e} induces from  MVSEC and generates car annotations with the help of the pre-trained YOLO-v3 \cite{redmon2018yolov3} for night scenarios. Sharing a similar motivation, \cite{tomy2022fusing} uses the pre-trained YOLO-v5 \cite{yolov5} to create the bounding boxes for objects from the DSEC dataset. There are also several other datasets targeting segmentation tasks, such as the DET dataset \cite{cheng2019det} which provides annotation for lane detection. However, these works focus more on 2D segmentation, with few works taking the temporal consistency and object motion into account. For autonomous driving, understanding object motions is crucial for driving safety. Therefore, we propose an RGB-Event dataset, called DSEC-MOD, to encourage further research on moving object detection.

\noindent\textbf{Event Processing:}
Event cameras have recently drawn great attention thanks to their asynchronous characteristic. As bio-inspired sensors, several event-by-event researches aim to develop novel spiking neural network \cite{kundu2021spike, wozniak2020deep,shrestha2018slayer} to process this novel sensor information \cite{lee2016training,rueckauer2017conversion}. However, there are few works dealing with RGB-Event inputs, especially for multi-modal sensor fusion.

From another perspective, events can be naturally considered as 4D inputs $(x, y, p, t)$, where $(x,y)$ stands for the spatial resolution as that of images, $p$ stands for the polarity, and $t$ stands for the temporal axis. Until nowadays, there is no conventional event representation for video applications as suggested in previous works \cite{sun2022event, gallego2020event}. In the literature, there are three popular representations: frame-like methods \cite{zhu2018ev, gehrig2020eklt, liu2018adaptive, Mondal_2021_ICCV} that focus more on dealing with $(x, y, p)$ information, time-surface methods \cite{sironi2018hats, manderscheid2019speed, zhou2018semi} that focus more on $(x, y, t)$ cues, and voxel-based methods \cite{tulyakov2021time, rebecq2019events, zhu2019unsupervised, bardow2016simultaneous} that deal with all dimensions at the same time $(x, y, p, t)$. While voxel-based methods can fully explore both space-time information, its 4D structure requires higher computational cost compared to frame-based and time-surface methods, which are not suitable for real-time applications such as autonomous driving. In this paper, we propose a novel event representation for moving object detection. We build upon frame-based approaches and additionally integrate the time information through a temporal multi-scale aggregation, 
resulting in a simple yet effective manner to explore spatio-temporal cues.

\noindent\textbf{RGB-Event Fusion:}
With the development of event cameras, RGB-Event multi-modal fusion has drawn increasing research attention. Some works adopt a pre-trained network to realize event-frame conversion. Then, several works directly process the generated video in the same way as dealing with RGB video to achieve the target application. A recent work \cite{Sun22eccv} applies knowledge distillation to guide the learning on videos generated from events. However, we argue that there is a significant difference between knowledge distillation and multi-modal fusion. Other works take RGB and events from the input side and realize features during the feature modeling. \cite{Tulyakov2022timlens++,messikommer2022multi} proposes a late-fusion design to integrate events into RGB streaming to guide the feature decoding. Meanwhile, \cite{tomy2022fusing} adopts a middle-fusion design through simple concatenation convolution. \cite{sun2022event} further leverages the transformer's attention to better guide the feature fusion. Sharing a similar idea, we also follow the middle-fusion design to fuse RGB and events. Different from \cite{tomy2022fusing, sun2022event} with uni-directional fusion design, as shown in Fig. \ref{Fig-intro}, we explicitly model both modality-specific features and shared representation. Furthermore, we introduce a cross-modal calibration module to attentively correct the noisy response in each modality before feature fusion. Different from \cite{sun2022event} which focuses on the channel axis, our attention module further attends to the spatial dimension, which plays an important role for object localization within a frame.

\section{Our Work}

\begin{figure*}[t]
\centering
\includegraphics[width=\linewidth]{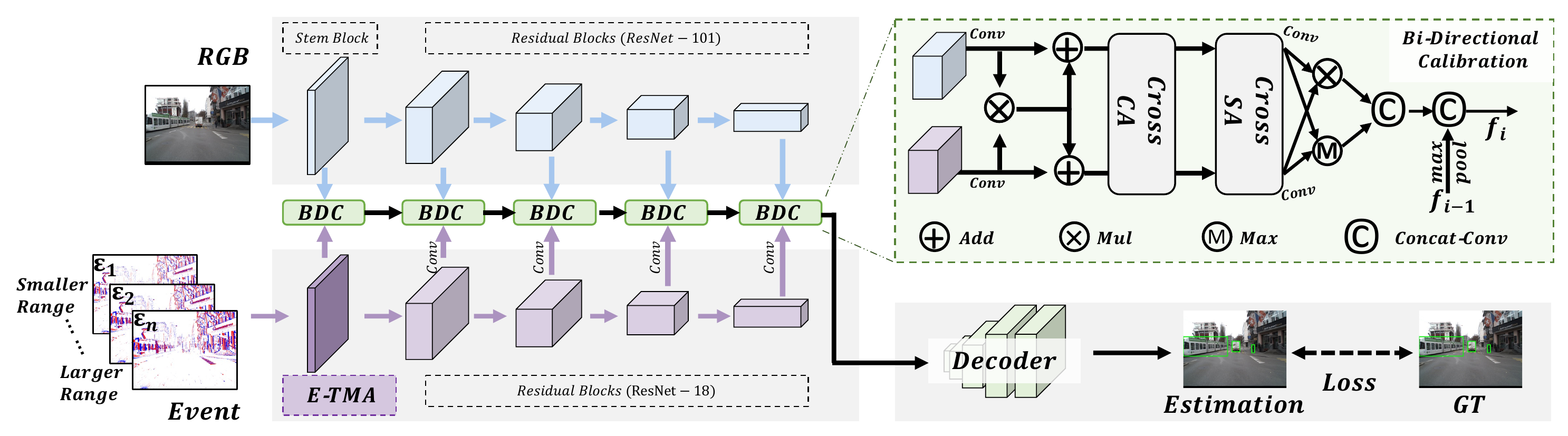}
\caption{\textbf{Architecture.} Our proposed network consists of an Event-based Temporal Multi-scale Aggregation (E-TMA see Sec. \ref{etma}) and an attentive fusion with Bi-Directional Calibration (BDC see Sec. \ref{bdc}). E-TMA aims to fully benefit from the temporal cues to improve the feature modeling, while BDC leverages cross-modal cues to firstly calibrate the noisy responses and attentively fuse them. The decoder and loss are adapted from \cite{li2020actions}.}
\label{Fig-RENet}
\vspace{-3mm}
\end{figure*}

Fig. \ref{Fig-RENet} presents the overall framework of our network RENet. For simplicity, we only illustrate the case for single-frame RGB-Event inputs. In practice, inputs use several frames from video as well as the associated events. 

Our network is composed of a tailored stemming layer to merge temporal multi-scale events (E-TMA see Sec. \ref{etma}), a dual encoder network to extract features, a bi-directional calibration module to firstly improve and then generate the shared representation (BDC see Sec. \ref{bdc}), and a decoder to output the bounding box of moving objects. Specifically, the multi-scale events are first fed into our proposed E-TMA to fully leverage the rich temporal motion cues.  Then, we adopt a dual but discrepant residual streaming design to extract semantic maps from each modality. To realize the multi-modal fusion, we attend to both the channel and spatial dimensions of each modality and then enable a cross-modal calibration. This process aims to neglect inherent single-modal noisy response and improve the feature modeling. Then, the enhanced multi-modal features together with the hierarchical features are fused in a coarse-to-fine manner. Finally, the encoded features are fed into the MOD detector \cite{li2020actions} to produce the bounding box.  The details will be introduced in the following sections.

\subsection{E-TMA: Event-based Temporal Multi-scale Aggregation
}
\label{etma}

\begin{figure}[t]
\centering
\includegraphics[width=\linewidth]{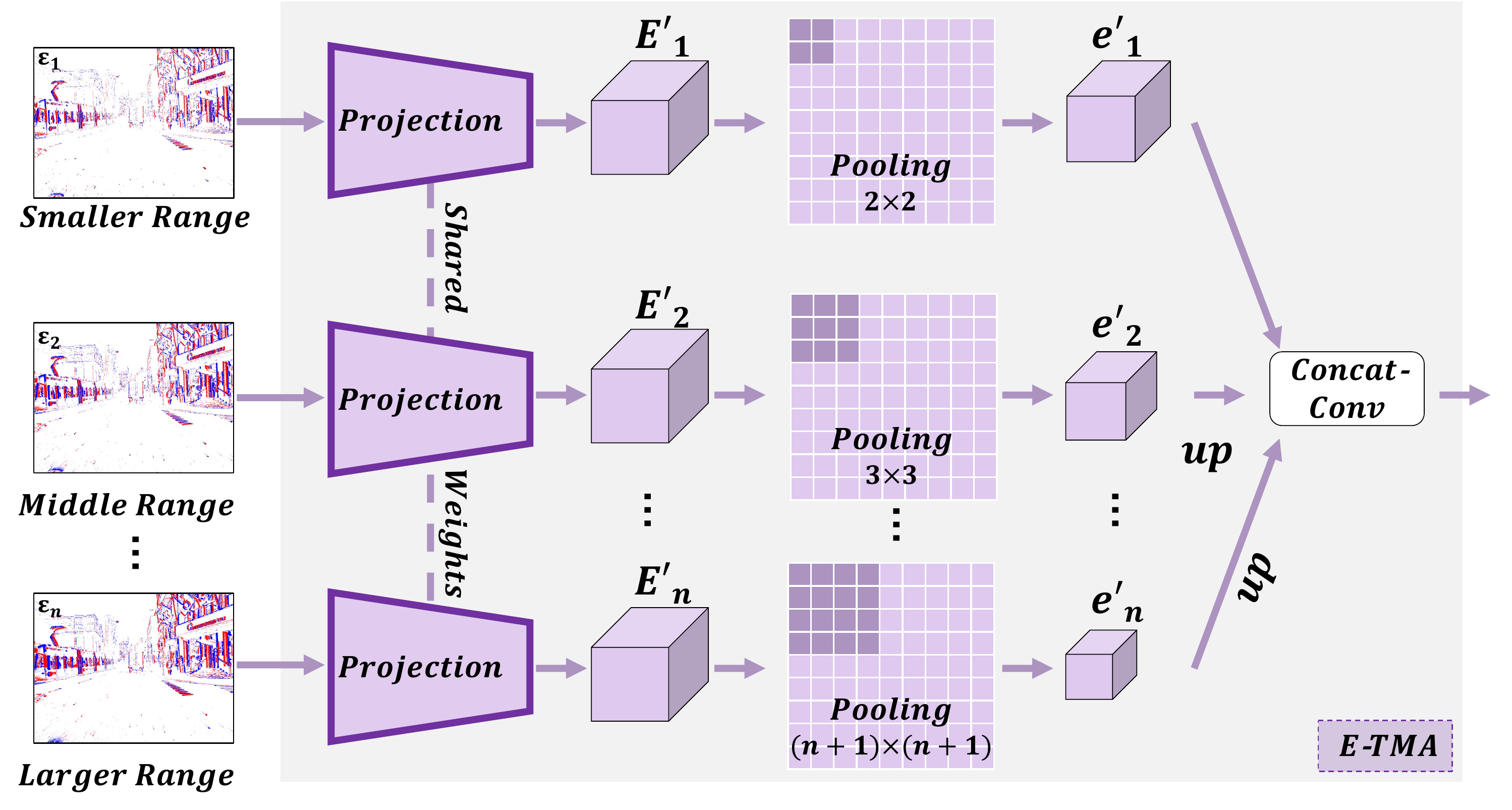}
\caption{\textbf{E-TMA: Event-based Temporal Multi-scale Aggregation.} Input events are first projected into a latent space. Then, different sizes of pooling are applied according to the event range. Finally, event features are merged together to learn the shared output with enhanced temporal consistency.}
\label{Fig-ETMA}
\vspace{-3mm}
\end{figure}

We observe from Fig. \ref{Fig-ETMA} that objects can slightly move from one position to another, i.e., there are more events on object boundaries while the temporal range of events becomes lager. However, the position changes are not significant nor sudden. Inspired by this observation, we design an aggregation module to establish the temporal relationship between multi-range event features $E'_i$. The aggregation must have two properties. First, it must maintain the homogeneous semantic cues across different ranges since the overall environment remains the same. Second, it must also be sensitive to spatially and temporally moving positions. Most existing works often process events within a pre-defined range, neglecting rich dynamic characteristics of event cameras.

To address this issue, we propose a multi-scale event aggregation module. Here, the term \textbf{Scale} does \textbf{Not} stand for the spatial resolution as conventionally used in image-based tasks. Instead, it stands for the different \textbf{Temporal} ranges widely used in the signal processing domain. Mathematically, during a specific temporal pool $T = \{ T_i | i \in N* \}$ under the rule that $ T_i \leq T_{i+k} $ with $ k \in N $, the groups of events $ \varepsilon_i $ are generated respectively. In our case, we take three scales of events as inputs, i.e, the events obtained from the RGB frame exposure time ($\varepsilon_1$), the events obtained from an intermediate scale which is double the exposure time in our case ($\varepsilon_2$), and the events obtained during one RGB frame time ($\varepsilon_3$).
To deal with multi-scale events, we first project the input events into a latent feature space. The projection module, denoted as $\eta$,  is similar to the first VGG layer which is composed of a combination of Convolution, Batch Normalization, and Relu activation. Each scale of events is fed into the same projection module to generate the event feature map. In other words, the projection weights are shared for each scale event input. The projection step can be expressed as:
\begin{equation}
   E'_i = \eta (\varepsilon_i), i \in {1,2,3}.
\end{equation}

To model the objects moving within a local region, pooling operations with different window sizes are applied to the event features: features from long-range events are processed with coarser pooling, while features from small-range events are processed with finer pooling. Formally, let the pooling with different window sizes $pool_i = \{pool_1; pool_2; pool_3 | size(pool_1)< size(pool_2)<size(pool_3)\}$, we apply max pooling on top multi-scale event features $E'_i$:
\begin{equation}
   e'_i = pool_i(E'_i), i \in {1,2,3}.
\end{equation}

Note that due to the different sizes of the pooling kernels, the pooled features do not share the same resolution. To merge them together, upsampling is applied to smaller-shape features to adjust the resolution. Finally, all features are concatenated together and fed into a convolution to generate the event representation $f_e$. We have:
\begin{equation}
   f_e = Conv(Concat(e'_1; up(e'_2); up(e'_3))),
\end{equation}
where $up$ denotes the upsampling to match the resolution. The aggregated feature is later fed into the residual encoder \cite{he2016deep} for semantic modeling.

Our event modeling is significantly different from previous works. Several works simply take a pre-defined range of events as input \cite{tomy2022fusing, Tulyakov2022timlens++}, which cannot fully leverage the rich temporal cues. A recent concurrent work \cite{sun2022event} also proposes to deal with multi-scale events. However, they simply accumulate the polarities of multi-scale events from the input side. Despite its demonstrated effectiveness on image deblurring, this method fails to explicitly distinguish the object boundary and the background, yielding a noisy representation for object detection tasks.  Differently, we propose to merge the multi-scale events at the latent space with the help of pooling operations. The choice of pooling is mainly based on its spatial scale and rotation invariant property which fits our application extremely well, as discussed previously. We particularly leverage the max pooling operation to extract the most informative cues in the event features. The motivation is based on the intuition that the object boundary should provide a strong response to distinguish the traffic participant from the background. To the best of our knowledge, we are the first to introduce a learnable aggregation module to extract, preserve, and fuse the most informative event features across different temporal scales. We show in Table \ref{Tab-Fusion} the superiority of our proposed event representation approach over our counterparts.

\subsection{Discrepant Two-Streaming Encoder}

Most existing multi-model models apply dual encoders for feature extraction. These encoders do not share weight but share the same architecture. However, we observe that RGB and event maps are significantly different, i.e., the former (RGB) provides dense textual information at each pixel, while the latter (event) only provides sparse activation for the pixels with brightness changes. Moreover, through our empirical experiments, as shown in Table \ref{Tab-Fusion}, we observe that event-only streaming yields significantly lower performance compared to RGB-only streaming. This observation motivates us to design a discrepant two-streaming encoder for feature extraction as shown in Fig. \ref{Fig-RENet},  i.e., a deeper encoder for RGB processing with ResNet-101 and a shallower encoder for event processing with Resnet-18. In such a way, we distinguish RGB and event modalities that the event stream plays an assist and auxiliary role. This discrepant design can also help to reduce computational costs.

To realize the multi-modal fusion, RGB and event features must share the same size. However, due to the discrepant encoders, the channel size for our RGB and event features are not the same. Therefore, before realizing the feature fusion, we apply a projection on the encoded event features to match the channel size of RGB features.

\subsection{BDC: Bi-Directional Calibration}
\label{bdc}
RGB-Event encoder fusion is a relatively novel research topic with very few papers compared to other multi-modal fusions such as RGB-Depth and RGB-Point cloud. Previous RGB-Event works or simply merge multi-modal features through simple concatenation convolution \cite{tomy2022fusing}, or attend to the channel dimension with the help of transformer attention \cite{sun2022event}. However, these methods do not explicitly tackle the spatial cues which are essential for segmentation/detection tasks. A recent tentative on knowledge transfer \cite{Sun22eccv} has also shown great success for event processing. Nevertheless, we argue that there is a huge difference between multi-modal fusion and knowledge distillation where the latter cannot fully leverage the cross-modal features compared to the former. Therefore, we seek to design an attentive fusion that can leverage the informative cues from both spatial and channel axis during the encoding stage.

For simplicity, we take the fusion at the highest semantic level as an example. Formally, let the RGB feature $f_{R} \in \mathbb{R}^{C\times h \times w}$ and the projected event feature $f_{E} \in \mathbb{R}^{C\times h \times w}$. We first apply a transformation module based on $1\times 1$ convolution for activation. We obtain the activated feature maps $f_{r}$ and $f_{e}$ by:
\begin{equation}
    f_{r} = Conv_{1\times1}(f_R); \quad    f_{e} = Conv_{1\times1}(f_E).
\end{equation}

Then we attentively fuse RGB and event features in a coarse-to-fine manner. Specifically, we first
use a pixel-wise multiplication along with addition to coarsely enhance the most informative features in each modality.
Formally, the enhanced features $f'_{r}$ and $f'_{e}$ are computed by:
\begin{equation}
    f'_{r} = f_r \otimes f_e + f_r; \quad    f'_{e} = f_r \otimes f_e + f_e.
\end{equation}

Then we refine the features by explicitly attending to the channel and spatial axes in a separate and successive manner. For each axis, we enable a bi-directional calibration, i.e., we learn the features from one modality and apply it to the other. Then, we merge calibrated multi-modal features while maintaining the most informative features. For simplicity, we show the example with channel calibration. Let $CA$ be the channel attention module from \cite{woo2018cbam}. The cross-calibration along the channel axis can be formulated as:
\begin{equation}
f^{CA}_{r} = CA(f'_{e}) \otimes f'_r + f'_r; \quad f^{CA}_{e} = CA(f'_r) \otimes f'_e + f'_e.
\end{equation}

Sharing the same protocol, we further refine the features spatially. Let $SA$ be the spatial attention from \cite{woo2018cbam}, we obtain the final enhanced features $f^{enh}_{r}$ and $f^{enh}_{e}$ as follows:
\begin{equation}
f^{enh}_{r} = SA(f^{CA}_{e}) \otimes f^{CA}_{r} + f^{CA}_{r}; \quad f^{enh}_{e} = SA(f^{CA}_{r}) \otimes f^{CA}_{e} + f^{CA}_{e}.
\end{equation}

These designs aim to find the most truthful and confident channels/pixels within each modality-specific feature through the deep network. Moreover, cross-calibration can selectively calibrate the noisy features while maintaining the informative cues, yielding a simple yet efficient manner to boost the feature modeling with the help of complementary cues. Finally, the calibrated features are attentively merged through a convolution that learns the contribution weights from the most informative components to form the shared output:
\begin{equation}
f = Conv_{3\times 3}(Concat(f^{enh}_{r} \otimes f^{enh}_{e}; max(f^{enh}_{r}, f^{enh}_{e}))).
\end{equation}

Further, the fused output is integrated with hierarchical features, i.e., the output from the previous layer (if any).

\section{EXPERIMENTS}
\subsection{Our RGB-Event Moving Object Dataset}

To the best of our knowledge, DSEC is the largest RGB-Event dataset for autonomous driving. However, DSEC only proposes the depth ground truth, without object bounding box annotation. \cite{tomy2022fusing} proposes to generate object annotation with the pre-trained YOLO-V5 network. However, the used dataset is not available. Moreover, the generated annotation takes both statistics and dynamic objects into account. Therefore, to promote the research of moving object detection with events, we introduce a novel DSEC-MOD dataset.

Specifically, we first calibrate the multi-modal input. Following \cite{gehrig2021dsec}, we project the RGB frames to the event-based coordinates with the help of camera matrices. In this process, RGB and event maps have the same field of view and the same resolution.
Then, based on the calibrated RGB videos, we manually label the annotation for moving objects.
In our dataset, 8 object classes are taken into account: car, truck, bus, train, pedestrian, cyclist, motorcyclist, and others.
Since we aim to build a challenging dataset with multiple moving objects in the scene, our sequences contain at least 3 different types of moving objects. In total, our DSEC-MOD dataset contains 16 sequences (13314 frames), with 11 sequences (10495 frames) for training and 5 other sequences (2819 frames) for testing. To the best of our knowledge, our moving object dataset is the largest of its kind.



\begin{table*}[t]
\caption{
\textbf{Comparison with SOTA fusion alternatives.}
0.5 and 0.2 stand for the thresholds of the intersection of union.}
\label{Tab-Fusion}
\begin{center}
\begin{NiceTabular}{l|l|c|c|c|c}

\hline

\hline
Model Type & Fusion Module & Pub. \& Year & F. mAP @0.5 (\%) & V. mAP @0.2 (\%) & V. mAP @0.5 (\%) \\
\hline

\hline
\multirow{4}{*}{Basic Setting} & 
RGB only & / & 30.75 & 12.82 & 4.97 \\
& Event only & / & 14.66 & 4.50 & 0.10 \\
& Early Fusion & / & 32.27 & 17.81 & 5.43 \\
& Late Fusion & / & 34.27 & 19.08 & 6.41 \\
\hline
\multirow{3}{*}{Self-Attention} 
& SENet \cite{hu2018squeeze} & CVPR'18 & 29.28 & 17.51 & 3.77 \\
& CBAM \cite{woo2018cbam} & ECCV'18 & 36.22 & 16.36 & 5.63 \\
& ECANet \cite{wang2020eca} & CVPR'20 & 34.49 & 18.79 & 6.73 \\

\hline
\multirow{3}{*}{Cross-Attention} &
SAGate \cite{chen2020bi} & ECCV'20 & 33.62 & 17.80 & 5.34 \\
& DCF \cite{ji2021calibrated} & CVPR'21 & 32.20 & 18.84 & 4.39 \\
& SPNet \cite{zhou2021specificity} & ICCV'21 & 32.70 & 14.71 & 6.57 \\
\hline
\multirow{3}{*}{RGB-Event} &
FPN-Fusion \cite{tomy2022fusing} & ICRA'22 & 32.28 & 14.97 & 5.74 \\
& EFNet \cite{sun2022event} & ECCV'22 & 35.33 & 18.05 & 6.30 \\
& \textbf{RENet} & \textbf{(Ours)} & \textbf{38.38} & \textbf{19.60} & \textbf{7.06} \\
\hline

\hline
\end{NiceTabular}
\end{center}
\vspace{-3mm}
\end{table*}

\subsection{Experimental Setup}

\noindent\textbf{Implementation Details:} We choose ResNet-101 and ResNet-18 as our backbones. The input frames are resized to $288\times 288$. We apply classical data augmentations such as photometric transformation, scale, and location jittering. We follow \cite{li2020actions} to use adam optimizer with the initial learning rate 5e-4, then decreasing by a factor of 10 at the 10th epochs and 15th epochs. The whole model is trained for 30 epochs.

\noindent\textbf{Evaluation Metrics:} We use frame mean average precision (F. mAP) and video mean average precision (V. mAP) for evaluation. Frame mAP evaluates the detection quality in a single frame, while video mAP also considers the linking of bounding boxes from a temporal perspective  \cite{weinzaepfel2015learning, gkioxari2015finding, kalogeiton2017action}.

\subsection{Comparison}

\noindent\textbf{Comparison with SOTA fusion alternatives:}
To the best of our knowledge, there are very few works on RGB-Event fusion during the encoder stage. Moreover, we are the first to conduct a study on moving object detection with RGB-Event input. Therefore, to purely and fairly analyze our performance, we replace our proposed modules with the state-of-the-art fusion alternatives from RGB-Event domain: FPN-Fusion \cite{tomy2022fusing}, EFNet \cite{sun2022event}; and from RGB-D domain: SAGate \cite{chen2020bi}, DCF \cite{ji2021calibrated}, SPNet \cite{zhou2021specificity}. We also compare with famous self-attention modules such as SENet \cite{hu2018squeeze}, ECAnet \cite{wang2020eca}, and CBAM \cite{woo2018cbam}. Specifically, we maintain the same multi-scale event input with our E-TMA (if with event input), same backbone, same decoder, loss, and same training settings as ours. The only difference is in the RGB-Event fusion module. Note that for all listed SOTA alternatives, we apply the attention modules through middle fusion.

As shown in Table \ref{Tab-Fusion}, compared to both self-attention and cross-modal attention works, our RENet outperforms its counterparts by a wide margin, validating our proposed fusion design. Specifically, compared to FPN-Fusion \cite{tomy2022fusing} which is based on middle fusion with concat-conv, our fusion is based on attention which can better deal with informative features across different modalities. Different from EFnet \cite{sun2022event} which attends only to the channel dimension, our fusion leverages both channel and spatial attention and realizes in addition a cross-calibration to improve the feature modeling.

\noindent\textbf{Comparison under different illumination conditions:} To better understand the contribution of event cameras, we analyze the performance gain under different illumination conditions. The detailed comparison can be found in Table \ref{Tab-Illumination}. It can be seen that with the complementary events, our proposed modules can significantly and consistently boost the baseline performance under different lighting conditions.

\begin{figure}[t]
\centering
\includegraphics[width=\linewidth]{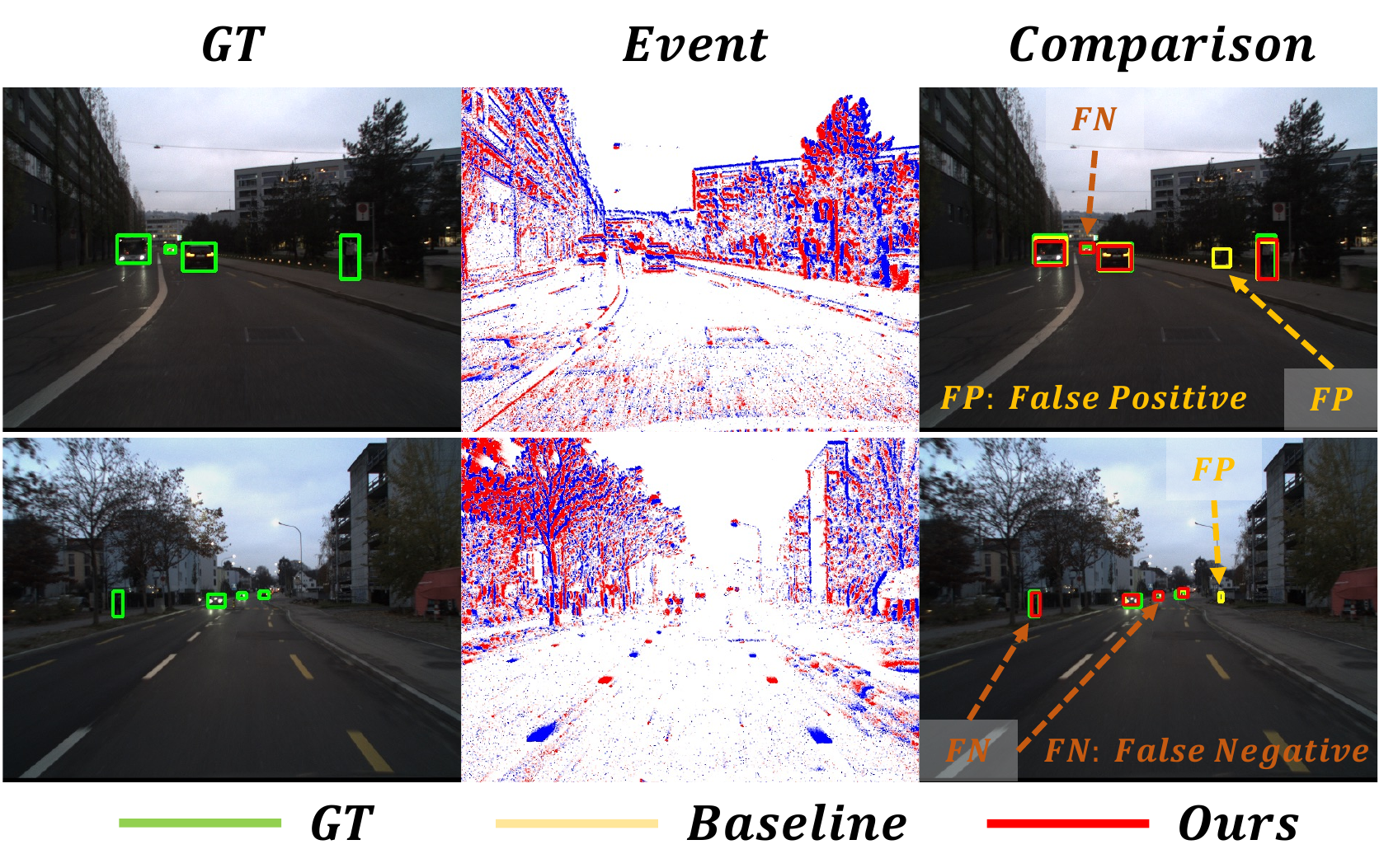}

\caption{\textbf{Qualitative comparison.} Our network generates more accurate detection with fewer failure cases. Best zoomed in.
}
\label{Fig-Vis}
\vspace{-3mm}
\end{figure}

\noindent\textbf{Qualitative Comparison:} Fig. \ref{Fig-Vis} depicts the qualitative comparison between our method and the RGB baseline. We can visualize the significant improvement in detecting objects with the help of events, especially in detecting small objects on the scene that are extremely challenging for RGB baseline. Moreover, when RGB baseline results in false positive due to the inferior lighting condition, our RGB-Event network can reason about more robust detection.


\begin{table}[t]
\caption{\textbf{Empirical comparison} under overall and specific illumination conditions.}
\label{Tab-Illumination}
\begin{center}
\begin{tabular}{l|l|c|c|c}
\hline

\hline
Illum. & Comp. & F. mAP@0.5 & V. mAP@0.2 & V. mAP@0.5 \\
\hline

\hline
\multirow{2}{*}{Overall} &RGB & 30.75 & 12.82 & 4.97  \\
 &+ Ours & 38.38 (\textbf{+7.63}) & 19.60 (\textbf{+6.78}) & 7.06 (\textbf{+2.09}) \\
\hline

\hline
\multirow{2}{*}{Day} &RGB & 42.08 & 16.81 & 9.01 \\
 &+ Ours & 45.58 (\textbf{+3.50}) & 16.84 (\textbf{+0.03}) & 11.15 (\textbf{+2.14}) \\
\hline

\hline
\multirow{2}{*}{Twilight} & RGB & 24.44 & 13.69 & 6.15 \\
 &+ Ours & 36.02 (\textbf{+11.58}) & 25.82 (\textbf{+12.13}) & 7.04 (\textbf{+0.89}) \\
\hline

\hline
\multirow{2}{*}{Night} & RGB & 17.44 & 15.79 & 0.76 \\
  & + Ours & 25.54 (\textbf{+8.10}) & 18.76 (\textbf{+2.97}) & 4.93 (\textbf{+4.17}) \\
\hline

\hline
\end{tabular}
\end{center}
\vspace{-3mm}
\end{table}

\subsection{Ablation Studies}
We also conduct ablation studies to analyze the contribution of each component. Specifically, we gradually add our proposed modules on top of the RGB baseline. It can be seen that each proposed method is helpful and can enable performance gain. We also replace our proposed multi-scale event aggregation with the recently proposed multi-scale accumulation \cite{sun2022event}. The main difference is that our methods aggregate multi-scale events at the semantic level in a coarse-to-fine manner, while \cite{sun2022event} simply concatenates multiple event frames from the input side. Empirically, we can see that the performance drops significantly, which demonstrates the superiority of our proposed module in modeling event data for (moving) object detection. 

\begin{table}[t]
\footnotesize
\setlength\tabcolsep{0.5pt}
\setlength\extrarowheight{1pt}
\begin{center}
\caption{\textbf{Ablation study on key components.} R-E stands for the RGB-Event baseline where we take a pre-defined range of events as input and features are merged through a simple concat-conv without any form of attention.}
\label{tab:3dvablation}
\begin{tabular}{m{.64cm} m{.64cm} m{.64cm} m{.64cm}  m{1cm} m{1.5cm} m{1.5cm} }
\hline

\hline

\# & R-E & CA & SA & E-TMA & Acc. \cite{sun2022event} & F. mAP@0.5 \\
\hline
1 &\checkmark & & & & & 34.25 \\
2 &\checkmark & & & \checkmark& & 35.56 \\
3 &\checkmark & \checkmark & & \checkmark& & 36.53 \\
4 &\checkmark & & \checkmark & \checkmark & & 37.75 \\
5 &\checkmark & \checkmark & \checkmark & & \checkmark& 36.34 \\
6 &\checkmark & \checkmark & \checkmark & \checkmark& & \textbf{38.38} \\

\hline

\hline
\end{tabular}
\end{center}
\vspace{-3mm}
\end{table}

\section{CONCLUSION}
In this paper, we proposed a novel fusion architecture for moving object detection with RGB-Event inputs. Different from previous works based on single-scale event inputs, we introduce a multi-scale events aggregation to explicitly leverage the rich temporal cues from the asynchronous event sensor. Furthermore, we gradually fuse the heterogeneous RGB and event features in a coarse-to-fine manner, yielding a simple yet efficient cross-calibration mechanism with the help of different forms of attention. Finally, we propose a novel DSEC-MOD dataset to encourage further research on RGB-Event fusion for moving object detection. Extensive comparisons of our proposed dataset validate the effectiveness and robustness of our approach compared to the state-of-the-art fusion alternatives. In our future work, we will test our methods on more recent approaches. We will also collect a real-world RGB-Event dataset for autonomous driving.










\bibliographystyle{IEEEtran}
\bibliography{IEEEexample}

\begin{thebibliography}{10}
\providecommand{\url}[1]{#1}
\csname url@rmstyle\endcsname
\providecommand{\newblock}{\relax}
\providecommand{\bibinfo}[2]{#2}
\providecommand\BIBentrySTDinterwordspacing{\spaceskip=0pt\relax}
\providecommand\BIBentryALTinterwordstretchfactor{4}
\providecommand\BIBentryALTinterwordspacing{\spaceskip=\fontdimen2\font plus
\BIBentryALTinterwordstretchfactor\fontdimen3\font minus
  \fontdimen4\font\relax}
\providecommand\BIBforeignlanguage[2]{{%
\expandafter\ifx\csname l@#1\endcsname\relax
\typeout{** WARNING: IEEEtran.bst: No hyphenation pattern has been}%
\typeout{** loaded for the language `#1'. Using the pattern for}%
\typeout{** the default language instead.}%
\else
\language=\csname l@#1\endcsname
\fi
#2}}

\bibitem{siam2018modnet}
M.~Siam, H.~Mahgoub, M.~Zahran, S.~Yogamani, M.~Jagersand, and A.~El-Sallab,
  ``Modnet: Motion and appearance based moving object detection network for
  autonomous driving,'' in \emph{International Conference on Intelligent
  Transportation Systems (ITSC)}, 2018.

\bibitem{rashed2019fusemodnet}
H.~Rashed, M.~Ramzy, V.~Vaquero, A.~El~Sallab, G.~Sistu, and S.~Yogamani,
  ``Fusemodnet: Real-time camera and lidar based moving object detection for
  robust low-light autonomous driving,'' in \emph{Proceedings of the IEEE/CVF
  International Conference on Computer Vision Workshops (ICCVW)}, 2019.

\bibitem{Baur_2021_ICCV}
S.~A. Baur, D.~J. Emmerichs, F.~Moosmann, P.~Pinggera, B.~Ommer, and A.~Geiger,
  ``Slim: Self-supervised lidar scene flow and motion segmentation,'' in
  \emph{Proceedings of the IEEE/CVF International Conference on Computer Vision
  (ICCV)}, 2021.

\bibitem{mackin2018study}
A.~Mackin, F.~Zhang, and D.~R. Bull, ``A study of high frame rate video
  formats,'' \emph{IEEE Transactions on Multimedia (TMM)}, vol.~21, no.~6, pp.
  1499--1512, 2018.

\bibitem{madhusudana2021st}
P.~C. Madhusudana, N.~Birkbeck, Y.~Wang, B.~Adsumilli, and A.~C. Bovik,
  ``St-greed: Space-time generalized entropic differences for frame rate
  dependent video quality prediction,'' \emph{IEEE Transactions on Image
  Processing (TIP)}, vol.~30, pp. 7446--7457, 2021.

\bibitem{sun2022coarse}
G.~Sun, Y.~Liu, H.~Ding, T.~Probst, and L.~Van~Gool, ``Coarse-to-fine feature
  mining for video semantic segmentation,'' in \emph{Proceedings of the
  IEEE/CVF Conference on Computer Vision and Pattern Recognition (CVPR)}, 2022.

\bibitem{chen2020event}
G.~Chen, H.~Cao, J.~Conradt, H.~Tang, F.~Rohrbein, and A.~Knoll, ``Event-based
  neuromorphic vision for autonomous driving: A paradigm shift for bio-inspired
  visual sensing and perception,'' \emph{IEEE Signal Processing Magazine
  (SPM)}, vol.~37, no.~4, pp. 34--49, 2020.

\bibitem{li2019event}
J.~Li, S.~Dong, Z.~Yu, Y.~Tian, and T.~Huang, ``Event-based vision enhanced: A
  joint detection framework in autonomous driving,'' in \emph{2019 IEEE
  International Conference on Multimedia and Expo (ICME)}, 2019.

\bibitem{maqueda2018event}
A.~I. Maqueda, A.~Loquercio, G.~Gallego, N.~Garc{\'\i}a, and D.~Scaramuzza,
  ``Event-based vision meets deep learning on steering prediction for
  self-driving cars,'' in \emph{Proceedings of the IEEE Conference on Computer
  Vision and Pattern Recognition (CVPR)}, 2018.

\bibitem{zhen2019estimating}
W.~Zhen and S.~Scherer, ``Estimating the localizability in tunnel-like
  environments using lidar and uwb,'' in \emph{International Conference on
  Robotics and Automation (ICRA)}, 2019.

\bibitem{wu2021modality}
Z.~Wu, G.~Allibert, C.~Stolz, C.~Ma, and C.~Demonceaux, ``Modality-guided
  subnetwork for salient object detection,'' in \emph{International Conference
  on 3D Vision (3DV)}, 2021.

\bibitem{wu2022robust}
Z.~Wu, S.~Gobichettipalayam, B.~Tamadazte, G.~Allibert, D.~P. Paudel, and
  C.~Demonceaux, ``Robust rgb-d fusion for saliency detection,''
  \emph{International Conference on 3D Vision (3DV)}, 2022.

\bibitem{lin2022r}
J.~Lin and F.~Zhang, ``R3 live: A robust, real-time, rgb-colored,
  lidar-inertial-visual tightly-coupled state estimation and mapping package,''
  in \emph{International Conference on Robotics and Automation (ICRA)}, 2022.

\bibitem{Messikommer2022bridging}
N.~Messikommer, D.~Gehrig, M.~Gehrig, and D.~Scaramuzza, ``Bridging the gap
  between events and frames through unsupervised domain adaptation,''
  \emph{IEEE Robotics and Automation Letters (RAL)}, vol.~7, no.~2, pp.
  3515--3522, 2022.

\bibitem{tomy2022fusing}
A.~Tomy, A.~Paigwar, K.~S. Mann, A.~Renzaglia, and C.~Laugier, ``Fusing
  event-based and rgb camera for robust object detection in adverse
  conditions,'' in \emph{International Conference on Robotics and Automation
  (ICRA)}, 2022.

\bibitem{Tulyakov2022timlens++}
S.~Tulyakov, A.~Bochicchio, D.~Gehrig, S.~Georgoulis, Y.~Li, and D.~Scaramuzza,
  ``Time lens++: Event-based frame interpolation with parametric non-linear
  flow and multi-scale fusion,'' in \emph{Proceedings of the IEEE/CVF
  Conference on Computer Vision and Pattern Recognition (CVPR)}, 2022.

\bibitem{gehrig2021raft}
M.~Gehrig, M.~Millh{\"a}usler, D.~Gehrig, and D.~Scaramuzza, ``E-raft: Dense
  optical flow from event cameras,'' in \emph{International Conference on 3D
  Vision (3DV)}, 2021.

\bibitem{sun2022event}
L.~Sun, C.~Sakaridis, J.~Liang, Q.~Jiang, K.~Yang, P.~Sun, Y.~Ye, K.~Wang, and
  L.~Van~Gool, ``Event-based fusion for motion deblurring with cross-modal
  attention,'' in \emph{European Conference on Computer Vision (ECCV)}, 2022.

\bibitem{tulyakov2021time}
S.~Tulyakov, D.~Gehrig, S.~Georgoulis, J.~Erbach, M.~Gehrig, Y.~Li, and
  D.~Scaramuzza, ``Time lens: Event-based video frame interpolation,'' in
  \emph{Proceedings of the IEEE/CVF Conference on Computer Vision and Pattern
  Recognition (CVPR)}, 2021.

\bibitem{tulyakov2022time}
S.~Tulyakov, A.~Bochicchio, D.~Gehrig, S.~Georgoulis, Y.~Li, and D.~Scaramuzza,
  ``Time lens++: Event-based frame interpolation with parametric non-linear
  flow and multi-scale fusion,'' in \emph{Proceedings of the IEEE/CVF
  Conference on Computer Vision and Pattern Recognition (CVPR)}, 2022.

\bibitem{gehrig2021dsec}
M.~Gehrig, W.~Aarents, D.~Gehrig, and D.~Scaramuzza, ``Dsec: A stereo event
  camera dataset for driving scenarios,'' \emph{IEEE Robotics and Automation
  Letters (RAL)}, vol.~6, no.~3, pp. 4947--4954, 2021.

\bibitem{hu2021v2e}
Y.~Hu, S.-C. Liu, and T.~Delbruck, ``v2e: From video frames to realistic dvs
  events,'' in \emph{Proceedings of the IEEE/CVF Conference on Computer Vision
  and Pattern Recognition (CVPR)}, 2021.

\bibitem{hidalgo2020learning}
J.~Hidalgo-Carri{\'o}, D.~Gehrig, and D.~Scaramuzza, ``Learning monocular dense
  depth from events,'' in \emph{International Conference on 3D Vision (3DV)},
  2020.

\bibitem{zhu2018multivehicle}
A.~Z. Zhu, D.~Thakur, T.~{\"O}zaslan, B.~Pfrommer, V.~Kumar, and K.~Daniilidis,
  ``The multivehicle stereo event camera dataset: An event camera dataset for
  3d perception,'' \emph{IEEE Robotics and Automation Letters (RAL)}, vol.~3,
  no.~3, pp. 2032--2039, 2018.

\bibitem{redmon2018yolov3}
J.~Redmon and A.~Farhadi, ``Yolov3: An incremental improvement,'' \emph{arXiv
  preprint arXiv:1804.02767}, 2018.

\bibitem{yolov5}
Ultralytics, ``Yolov5,'' \url{https://github.com/ultralytics/yolov5}.

\bibitem{cheng2019det}
W.~Cheng, H.~Luo, W.~Yang, L.~Yu, S.~Chen, and W.~Li, ``Det: A high-resolution
  dvs dataset for lane extraction,'' in \emph{Proceedings of the IEEE/CVF
  Conference on Computer Vision and Pattern Recognition Workshops (CVPRW)},
  2019.

\bibitem{kundu2021spike}
S.~Kundu, G.~Datta, M.~Pedram, and P.~A. Beerel, ``Spike-thrift: Towards
  energy-efficient deep spiking neural networks by limiting spiking activity
  via attention-guided compression,'' in \emph{Proceedings of the IEEE/CVF
  Winter Conference on Applications of Computer Vision (WACV)}, 2021.

\bibitem{wozniak2020deep}
S.~Wo{\'z}niak, A.~Pantazi, T.~Bohnstingl, and E.~Eleftheriou, ``Deep learning
  incorporating biologically inspired neural dynamics and in-memory
  computing,'' \emph{Nature Machine Intelligence (Nat. Mach. Intell)}, vol.~2,
  no.~6, pp. 325--336, 2020.

\bibitem{shrestha2018slayer}
S.~B. Shrestha and G.~Orchard, ``Slayer: Spike layer error reassignment in
  time,'' \emph{Advances in Neural Information Processing Systems (NIPS)},
  2018.

\bibitem{lee2016training}
J.~H. Lee, T.~Delbruck, and M.~Pfeiffer, ``Training deep spiking neural
  networks using backpropagation,'' \emph{Frontiers in Neuroscience (Front.
  Neurosci.)}, vol.~10, p. 508, 2016.

\bibitem{rueckauer2017conversion}
B.~Rueckauer, I.-A. Lungu, Y.~Hu, M.~Pfeiffer, and S.-C. Liu, ``Conversion of
  continuous-valued deep networks to efficient event-driven networks for image
  classification,'' \emph{Frontiers in Neuroscience (Front. Neurosci.)},
  vol.~11, p. 682, 2017.

\bibitem{gallego2020event}
G.~Gallego, T.~Delbr{\"u}ck, G.~Orchard, C.~Bartolozzi, B.~Taba, A.~Censi,
  S.~Leutenegger, A.~J. Davison, J.~Conradt, K.~Daniilidis, \emph{et~al.},
  ``Event-based vision: A survey,'' \emph{IEEE Transactions on Pattern Analysis
  and Machine Intelligence (TPAMI)}, vol.~44, no.~1, pp. 154--180, 2020.

\bibitem{zhu2018ev}
A.~Zhu, L.~Yuan, K.~Chaney, and K.~Daniilidis, ``Ev-flownet: Self-supervised
  optical flow estimation for event-based cameras,'' in \emph{Robotics: Science
  and Systems (RSS)}, 2018.

\bibitem{gehrig2020eklt}
D.~Gehrig, H.~Rebecq, G.~Gallego, and D.~Scaramuzza, ``Eklt: Asynchronous
  photometric feature tracking using events and frames,'' \emph{International
  Journal of Computer Vision (IJCV)}, vol. 128, no.~3, pp. 601--618, 2020.

\bibitem{liu2018adaptive}
M.~Liu and T.~Delbruck, ``Adaptive time-slice block-matching optical flow
  algorithm for dynamic vision sensors,'' in \emph{British Machine Vision
  Conference (BMVC)}, 2018.

\bibitem{Mondal_2021_ICCV}
A.~Mondal, S.~R, J.~H. Giraldo, T.~Bouwmans, and A.~S. Chowdhury, ``Moving
  object detection for event-based vision using graph spectral clustering,'' in
  \emph{Proceedings of the IEEE/CVF International Conference on Computer Vision
  Workshops (ICCVW)}, 2021.

\bibitem{sironi2018hats}
A.~Sironi, M.~Brambilla, N.~Bourdis, X.~Lagorce, and R.~Benosman, ``Hats:
  Histograms of averaged time surfaces for robust event-based object
  classification,'' in \emph{Proceedings of the IEEE Conference on Computer
  Vision and Pattern Recognition (CVPR)}, 2018.

\bibitem{manderscheid2019speed}
J.~Manderscheid, A.~Sironi, N.~Bourdis, D.~Migliore, and V.~Lepetit, ``Speed
  invariant time surface for learning to detect corner points with event-based
  cameras,'' in \emph{Proceedings of the IEEE/CVF Conference on Computer Vision
  and Pattern Recognition (CVPR)}, 2019.

\bibitem{zhou2018semi}
Y.~Zhou, G.~Gallego, H.~Rebecq, L.~Kneip, H.~Li, and D.~Scaramuzza,
  ``Semi-dense 3d reconstruction with a stereo event camera,'' in
  \emph{European Conference on Computer Vision (ECCV)}, 2018.

\bibitem{rebecq2019events}
H.~Rebecq, R.~Ranftl, V.~Koltun, and D.~Scaramuzza, ``Events-to-video: Bringing
  modern computer vision to event cameras,'' in \emph{Proceedings of the
  IEEE/CVF Conference on Computer Vision and Pattern Recognition (CVPR)}, 2019.

\bibitem{zhu2019unsupervised}
A.~Z. Zhu, L.~Yuan, K.~Chaney, and K.~Daniilidis, ``Unsupervised event-based
  learning of optical flow, depth, and egomotion,'' in \emph{Proceedings of the
  IEEE/CVF Conference on Computer Vision and Pattern Recognition (CVPR)}, 2019.

\bibitem{bardow2016simultaneous}
P.~Bardow, A.~J. Davison, and S.~Leutenegger, ``Simultaneous optical flow and
  intensity estimation from an event camera,'' in \emph{Proceedings of the IEEE
  Conference on Computer Vision and Pattern Recognition (CVPR)}, 2016.

\bibitem{Sun22eccv}
S.~Zhaoning, M.~Nico, G.~Daniel, and S.~Davide, ``Ess: Learning event-based
  semantic segmentation from still images,'' \emph{European Conference on
  Computer Vision (ECCV)}, 2022.

\bibitem{messikommer2022multi}
N.~Messikommer, S.~Georgoulis, D.~Gehrig, S.~Tulyakov, J.~Erbach,
  A.~Bochicchio, Y.~Li, and D.~Scaramuzza, ``Multi-bracket high dynamic range
  imaging with event cameras,'' in \emph{Proceedings of the IEEE/CVF Conference
  on Computer Vision and Pattern Recognition (CVPR)}, 2022.

\bibitem{li2020actions}
Y.~Li, Z.~Wang, L.~Wang, and G.~Wu, ``Actions as moving points,'' in
  \emph{European Conference on Computer Vision (ECCV)}, 2020.

\bibitem{he2016deep}
K.~He, X.~Zhang, S.~Ren, and J.~Sun, ``Deep residual learning for image
  recognition,'' in \emph{Proceedings of the IEEE Conference on Computer Vision
  and Pattern Recognition (CVPR)}, 2016.

\bibitem{woo2018cbam}
S.~Woo, J.~Park, J.-Y. Lee, and I.~S. Kweon, ``Cbam: Convolutional block
  attention module,'' in \emph{European Conference on Computer Vision (ECCV)},
  2018.

\bibitem{hu2018squeeze}
J.~Hu, L.~Shen, and G.~Sun, ``Squeeze-and-excitation networks,'' in
  \emph{Proceedings of the IEEE Conference on Computer Vision and Pattern
  Recognition (CVPR)}, 2018.

\bibitem{wang2020eca}
W.~Qilong, W.~Banggu, Z.~Pengfei, L.~Peihua, Z.~Wangmeng, and H.~Qinghua,
  ``{ECA-Net}: Efficient channel attention for deep convolutional neural
  networks,'' in \emph{Proceedings of the IEEE/CVF Conference on Computer
  Vision and Pattern Recognition (CVPR)}, 2020.

\bibitem{chen2020bi}
X.~Chen, K.-Y. Lin, J.~Wang, W.~Wu, C.~Qian, H.~Li, and G.~Zeng,
  ``Bi-directional cross-modality feature propagation with
  separation-and-aggregation gate for rgb-d semantic segmentation,'' in
  \emph{European Conference on Computer Vision (ECCV)}, 2020.

\bibitem{ji2021calibrated}
W.~Ji, J.~Li, S.~Yu, M.~Zhang, Y.~Piao, S.~Yao, Q.~Bi, K.~Ma, Y.~Zheng, H.~Lu,
  \emph{et~al.}, ``Calibrated rgb-d salient object detection,'' in
  \emph{Proceedings of the IEEE/CVF Conference on Computer Vision and Pattern
  Recognition (CVPR)}, 2021.

\bibitem{zhou2021specificity}
T.~Zhou, H.~Fu, G.~Chen, Y.~Zhou, D.-P. Fan, and L.~Shao,
  ``Specificity-preserving rgb-d saliency detection,'' in \emph{Proceedings of
  the IEEE/CVF International Conference on Computer Vision (ICCV)}, 2021.

\bibitem{weinzaepfel2015learning}
P.~Weinzaepfel, Z.~Harchaoui, and C.~Schmid, ``Learning to track for
  spatio-temporal action localization,'' in \emph{Proceedings of the IEEE
  International Conference on Computer Vision (ICCV)}, 2015.

\bibitem{gkioxari2015finding}
G.~Gkioxari and J.~Malik, ``Finding action tubes,'' in \emph{Proceedings of the
  IEEE Conference on Computer Vision and Pattern Recognition (CVPR)}, 2015.

\bibitem{kalogeiton2017action}
V.~Kalogeiton, P.~Weinzaepfel, V.~Ferrari, and C.~Schmid, ``Action tubelet
  detector for spatio-temporal action localization,'' in \emph{Proceedings of
  the IEEE International Conference on Computer Vision (ICCV)}, 2017.

\end{thebibliography}

\end{document}